\documentclass[letterpaper, 10 pt, conference]{ieeeconf}  % Comment this line out if you need a4paper

\IEEEoverridecommandlockouts                              % This command is only needed if 
\usepackage{graphics} % for pdf, bitmapped graphics files
\usepackage{epsfig} % for postscript graphics files
\usepackage{mathptmx} % assumes new font selection scheme installed
\usepackage{times} % assumes new font selection scheme installed
\usepackage{amsmath} % assumes amsmath package installed
\usepackage{amssymb}  % assumes amsmath package installed
\usepackage{xcolor}
\usepackage{wasysym}
\usepackage{textcomp}
\usepackage{physics}
\usepackage{algorithm}
\usepackage{cite}
\usepackage{algpseudocode}
\usepackage{xurl}
\usepackage{svg}
\newcommand{\argmax}{\mathop{\mathrm{argmax}}\limits}

\title{\LARGE \bf
Synergizing Morphological Computation and Generative Design:  \\  Automatic Synthesis of Tendon-Driven Grippers
}
\author{Kirill~D.~Zharkov$^{1}$, Mikhail~E.~Chaikovskii$^{1}$,  Yefim~V.~Osipov$^{1}$,  Rahaf Alshaowa$^{1}$, \\
Ivan~I.~Borisov$^{1}$, and Sergey~A.~Kolyubin$^{1}$
\thanks{$^{1}$Kirill~D.~Zharkov, Mikhail~E.~Chaikovskii, Yefim~V.~Osipov, Rahaf Alshaowa, Ivan~I.~Borisov, and Sergey~A.~Kolyubin are with the Biomechatronics and Energy-Efficient Robotics Lab, ITMO University, Saint Petersburg, Russia {\tt\small e-mail: \{borisovii, s.kolyubin\}@itmo.ru}}
}
\begin{document}

\maketitle
\thispagestyle{empty}
\pagestyle{empty}

%%%%%%%%%%%%%%%%%%%%%%%%
%%%%%%%%%%%%%%%%%%%%%%%%
\begin{abstract}

Robots' behavior and performance are determined both by hardware and software.  The design process of robotic systems is a complex journey that involves multiple phases. Throughout this process, the aim is to tackle various criteria simultaneously, even though they often contradict each other. The ultimate goal is to uncover the optimal solution that resolves these conflicting factors. Generative, computation or automatic designs are the paradigms aimed at accelerating the whole design process. 
%%%
Within this paper we propose a design methodology to generate linkage mechanisms for robots with morphological computation. We use a graph grammar and a heuristic search algorithm to create robot mechanism graphs that are converted into simulation models for testing the design output. To verify the design methodology we have applied it to a relatively simple quasi-static problem of object grasping. Designing a fully actuated gripper may seem simple, but we found a way to automatically design an underactuated tendon-driven gripper that can grasp a wide range of objects. This is possible because of its structure, not because of sophisticated planning or learning.
%%%
% The success of these grippers relies more on their mechanical design rather than complex control algorithms.  
To test the applicability of the proposed method in real engineering practice, we used it to create physical prototypes. Simulation results together with results of testing of physical prototypes are given at the end of the paper. The framework is open source and the link to GitHub is given in the paper.

\end{abstract}
%\vspace{0.5cm}

%%%%%%%%%%%%%%%%%%%%%%%%
%%%%%%%%%%%%%%%%%%%%%%%%
\section{Introduction}

Designing robots is a multiphase process aimed at solving a multi-criteria optimization problem to find the best possible detailed design. When it comes to creating something new, the possibilities are endless. Creator must consider factors like shape, mechanics, materials, sensors, controllers, and etc. to bring design to life. The solution space knows no bounds, making each design choice a crucial step towards bringing the vision to reality. When the designing process is manual, it is difficult to prove the optimal solution numerically because it relies on the designer's experience and engineering intuition~\cite{sun2021research}. 

%%% Computation design, automatic design, and generative design (GD) are intended to produce mechanically improved solutions based on robust and rigorous models of design conditions and performance criteria~\cite{pinskier2022bioinspiration}. GD aims to significantly reduce the time needed for synthesis and verification to find effective solutions.  GD has to shorten the time for synthesis and verification to find better solutions. The results should surpass the design solutions provided by experts, or at least be on a par with them.}

\begin{figure}[t] 
	\begin{center}
		\includegraphics[width=1\linewidth]{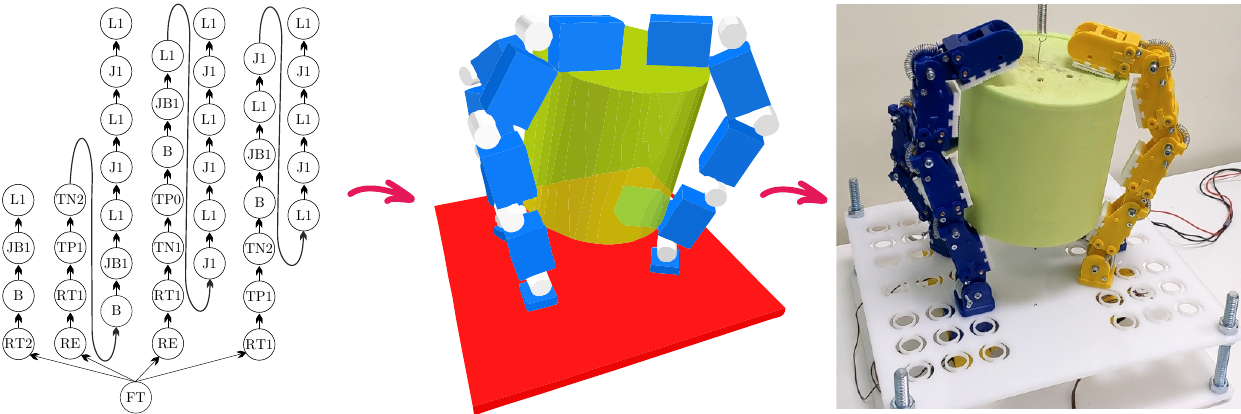}
		\caption{The paper proposes a generative design approach that is based on interaction between a graph generated by a heuristic algorithm (shown on the left) and a simulation model based on the graph (shown in the center). To verify generated designs and justify the proposed procedure, physical prototypes were built (on the right)} 
		\label{pic_1}
	\end{center}
\end{figure}

%A lot of effort has been put into utilizing finite element analysis to generate rigid and elastic bodies~\cite{transue2019generative}. Both types of bodies passively experience external loads and could be considered as static and quasi-static cases respectively; generative design of such systems is a mature technology and even a commercially available one~\cite{fusion}. On the other hand, the problem of generative design of \textit{dynamic systems} that transform motion of joints into complex links motion is more challenging and is still far away from being solved. 

\subsection{Contribution}

This research output is based on our previous studies on \textit{morphological computation} as \cite{borisov2021computational, borisov2022reconfigurable, nasonov2023computational}. As a step towards, we developed the ``rostok"\footnote{\url{https://github.com/aimclub/rostok}} framework for generative and interactive design of linkage mechanisms. The proposed framework is general enough and an advanced user can use it for a linkage synthesis for any robotic purposes: robot arms, legs, fingers etc. 

To verify applicability of the methodology for real engineering tasks, we have applied it to solve the task of computational design of underactuated tendon driven grippers, as a relatively simple quasi-static task. Fig.~\ref{pic_1} shows steps of the design process: (1) a heuristic algorithm generates a graph; (2) the graph is converted into a simulation model; (3) virtual experiments are conducted using the simulation model, and the results are evaluated using a reward function; (4) the reward value obtained is used to guide the exploration of the design space; (5) the final design options are verified by means of physical prototypes.

For the sake of clarity%explicitness
, the contribution of the paper consists of several items:
\begin{itemize}
    \item Combining \textit{morphological computation} paradigm with \textit{generative design}, the paper proposes an approach to the \textit{automatic generation} of linkage mechanisms with ``mechanical intelligence". The performance of these mechanisms is mainly determined by their mechanics, or \textit{morphology}, rather than sophisticated trajectory planning and control. 
    \item We successfully implemented the graph grammar approach to explore the design space of \textit{anthropomorphic underactuated tendon-driven grippers}. Grasping is done by means of gripper morphology, rather than sophisticated control strategies.
    \item To ensure the highest performance of generated grippers, \textit{physical prototypes} were constructed to validate both the solution and the overall methodology. This step aimed to assess the practicality and effectiveness of the methodology in real-world engineering applications. 
\end{itemize}

%========================
%========================
\subsection{Related Work}
Model-based design optimization approaches can be used  to efficiently synthesize linkage mechanisms together with needed control. Gradient-based together with global optimization algorithms are used to optimize geometry, mass distribution, or actuation within specific boundaries~\cite{thomaszewski2014computational, de2021control, chen2020underactuation}. Properly formalized optimization tasks allow to efficiently find suitable solutions, however model-based design optimization approaches strongly depend on user inputs.

As an alternative approach, the generative design needs minimum input user to create unique and unexpected solutions. However, the lack of the current simulation software the evolutionary design creates highly abstract structures that hardly can be physically reproduced~
%The work on co-design of robot morphology can be divided into two approaches. Voxel views of robots are convenient for representing soft robots
\cite{bhatia2021evolution,cheney2018scalable, ha2021fit2form}. %The design is modified by adding and removing voxels. As well as changing their physical characteristics. 

GD is intended to automatize co-optimization of mechanical structure and joints trajectories to satisfy poorly formalized tasks from a user. Generative adversarial networks (GANs)~\cite{hu2022modular}, graph neural networks (GNNs)~\cite{zhao2020robogrammar, wang2019neural}, and deep neural networks (DNNs)~\cite{gupta2022metamorph} are used to generate robots for a wide range of scenarios.
%In the case of solid bodies, the robot is represented in the form of an acyclic directed graph \cite{wang2018neural,gupta2021embodied}. In simple design is a kinematic tree. And adding limbs or removing limbs interacts with it. Allan Zhao et al. proposed a flexible method with interaction with topology \cite{zhao2020robogrammar}. 
The design of the robot is defined by a graph. Recursive grammar rules for changes are applied over the graph. The approach allows the search space to be modified by changing the set of rules. %Which allows us to set the space to design robots that can be manufactured.

%Model Predictive Path Integral Control (MPPI) to simulate robots is used in~~\cite{zhao2020robogrammar}. The MPPI is a variation of Model Predictive Path Integral Control when no differentiation is possible. Other works use parametrization of control policies. %Models are based on graph neural networks (GNN) to transform the robot \cite{wang2018neural, gupta2021embodied}.

A modular co-evolution strategy been implemented in~\cite{pathak19assemblies} where primitive agents dynamically self-assemble into complex bodies and learn to coordinate their actions. The approach outperformed static methods in simulated environments, showcasing improved adaptability to environmental changes and agent structure variations.

An innovative approach to robotic manipulator design, offering a streamlined pipeline that enables rapid creation and customization of manipulators with knitted, glove-like tactile sensors is given in~\cite{zlokapa2022integrated}. By applying modular components and predefined rules, engineers can quickly prototype manipulator designs, fine-tune their shape, and seamlessly integrate tactile sensing, with successful real-world applications demonstrated.

A novel underactuated compliant hand emulator, leveraging advanced simulation software and the adaptive synergy concept to simplify mechanical and control complexity, is introduced in~\cite{rocchi2016generic}. 

%Design automation for robots is addressed in~\cite{hu2022glso} through the introduction of Grammar-guided Latent Space Optimization, a framework that transforms the optimization process into a low-dimensional continuous problem, enabling significant improvements in sample efficiency through the utilization of algorithms like Bayesian Optimization, guided by graph grammar rules and robot world space features. %Importantly, the trained VAE can be reused to search for designs specialized for multiple tasks without retraining, as demonstrated in locomotion simulations, surpassing existing robot design automation methods.

%The other group of optimisation techniques are evolutionary ones. It finds various applications in the field of robotics \cite{doncieux2015evolutionary}. However, the existing solutions for anthropomorphic manipulators use a simplified parametric representation of the solution \cite{baressi2020path}. At the same time, it is promising to use directed acyclic graphs (DAGs) as a direct encoding of the trajectory of joints in genotype \cite{sotto2020study}.

\begin{figure*}[t] 
	\begin{center}
		\includegraphics[width=0.97\textwidth]{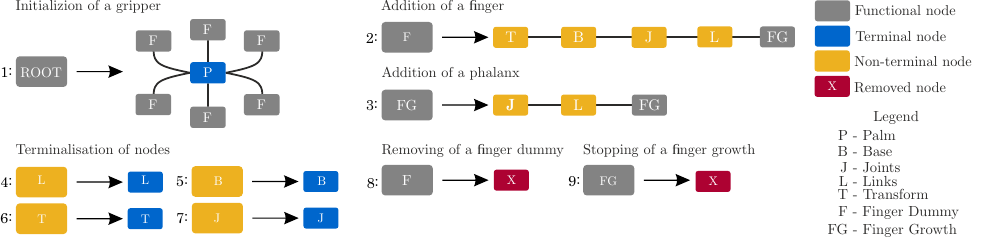}
		\caption{The vocabulary of rules: rule 1 initializes a palm P with finger dummies F, rule 2 adds a finger, replacing a finger dummy F with a group of nodes, rule 3 adds a phalanx instead of node FG, rules 4-7 terminate nodes, rule 8 removes a finger dummy F, and rule 9 stops a finger from growing} 
		\label{fig:pic_2}
	\end{center}
%\end{figure*}
%\begin{figure*}[t] 
	\begin{center}
		\includegraphics[width=0.97\textwidth]{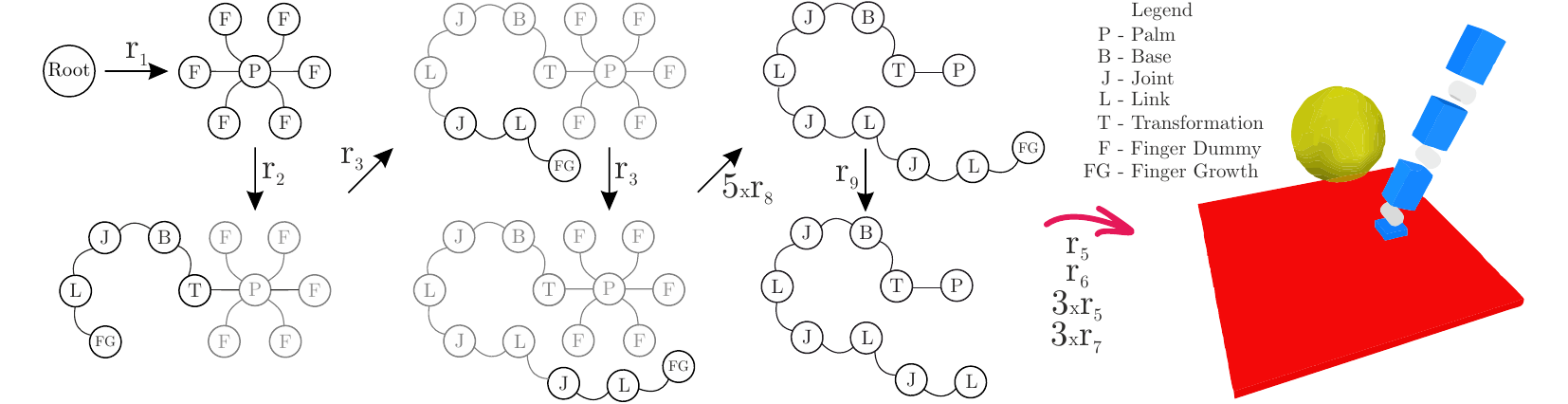}
		\caption{Instructive example of a sequence of rules applied to generate the simplest one-fingered gripper: rule $r_1$ transforms an initial node into a palm with 6 finger dummies F; rule $r_2$ transforms one of dummy F's into a finger with a base B, joint J, and an attachment for the next link FG; rule $r_3$ adds extra phalanges, while rules $r_8$ and $r_9$ eliminate non-terminal nodes such as F and FG, finally rules $r_5$, $r_6$, and $r_7$ terminate non-terminal nodes} 
		\label{fig:pic_3}
	\end{center}
\end{figure*}

%%%%%%%%%%%%%%%%%%%%%%%%
%%%%%%%%%%%%%%%%%%%%%%%%
\section{Methods}

% Currently, the applied algorithms support only generation of open chain mechanisms
In our approach the GD of dynamic systems can be decoupled into three sub-tasks: (1) \textit{generation of the mechanism morphology}, (2) \textit{optimisation of its kinematic and dynamics parameters}, and (3) \textit{optimisation of control efforts}. Within this paper, the main focus is on the morphology of an underactuated tendon-driven griper, i.e. how many links and how they are jointed to form a mechanism, such that the gripper can grasp objects blindly without object-shape-based trajectory planning or behavior cloning. The only information regarding an object to be grasped is the position of its geometrical center expressed in gripper's palm frame. 

%The paper focuses on a problem of synthesis of planar open chain underactuated tendon-driven linkage mechanisms.  

\subsection{Graph grammar}

%GD implies the usage of algorithms in the process of design creation.
Our approach requires a suitable way to represent different structures and to use search algorithms. For the ``rostok"  we chose to encode mechanisms in the form of graphs. The graph representation is frequently used for robot design~\cite{zhao2020robogrammar, wang2018neural, Sims_1994}. In a graph, nodes typically symbolize components of mechanisms while the edges define their connections. 

%Only a subset of possible graphs constructed from the 
% Using graph representation we convert the GD task to the problem of searching for 

To effectively use the graph representation of designs, we must limit the search space to the graphs that represent mechanisms that can be manufactured, i.e., satisfy the conditions of existence. We apply the graph grammar for two reasons: limit the graph space and guide the search algorithm. 

The graph grammar approach to graph synthesis requires two essential parts: \textit{set of nodes} and \textit{vocabulary of production rules} (Fig.~\ref{fig:pic_2}). The set of nodes includes all possible node types that can appear in the graphs, however, several nodes of the same type can occur in a graph. Within the paper we limited a range of nodes to the minimal needed to generated planar open-chain linkages with revolute joints. 

The production rules determine the possible transformations of a graph. Each rule mutates the current graph into a new one. As a result, in the %graph 
grammar scheme, the graph space is limited to the set of graphs that can be built by applying a consequence of rules. User can design the rule vocabulary and set of nodes in such a way that all graphs represent either the mechanism that can be assembled or some intermediate state that can be mutated to the mechanism representation via rule application. %Therefore, production rules are the convenient way to look over the design space. 
The node alphabet and rules depend on the specific task of GD and the design space must be explored. 

\subsection{Reward function}
We use the graph grammar to turn GD tasks into graph optimization problems. The optimization problem in graph space requires a mechanism of graph evaluation and comparison. This idea can be summarized into a \textit{reward function} that can evaluate each proposed design. Through the reward function one tells the algorithm what is actually considered to be a good mechanism. The design of the reward function is the part of the GD permitting the vast space for variation and requires the most creativity. 

In our paradigm of the GD, the algorithm has to \textit{generate graphs} and\textit{ calculate reward }to evaluate the generated solution. Graph grammar approach is especially effective for building mechanism morphology, because it can restrict the graph space to only feasible mechanisms. In ``rostok" we use graph grammar to search the morphology space. %Other design parts can be realized in various ways. 
User can optimize robot parts by using global optimization algorithms or by incorporating it into the production rules. The same is true for control optimization tasks; they can be performed independently or added to the grammar rules, e.g. torques limits can be set directly to the graph of an actuated joints. %In fully actuated open kinematic chains, torque values can be set directly to the graph. Alternatively, the graph can be used to set bounds for the control optimization algorithm. 

%For example, in the ``Robogrammar" \cite{zhao2020robogrammar} project, the parameters of optimization were incorporated into the grammar rules (the node alphabet included parts with various determined parameters), while the control optimization was performed independently for each built design. 

The ``rostok" pipeline assumes the usage of the mechanism \textit{simulations} as a main part of the reward calculation. Therefore, one needs to build an \textit{automatic way to convert the graph into the simulation models} with the desired setup. In our experience, it is the hardest part of the process, since simulation models must satisfy \textit{verification} and \textit{validation} to minimize \textit{reality gap}. %We need to make the simulation builder adaptable for each GD task. 
%Our goal is to simplify the automation process of the simulation. 
%In the tested applications we utilises the PyChrono multiphysical modeling engine, the Python version of the Project Chrono C++ modeling library [6], for multi-body simulation of the design mechanisms.
The framework utilizes the PyChrono multiphysical modeling engine, the Python version of the Project Chrono C++ modeling library~\cite{tasora2016chrono} for multi-body simulation of the design mechanisms.

As the reward calculation is simulation-based, our evaluation is limited to the graphs that depict the final robot design. However, it's important to note that the graph space encompasses the intermediate states as well. Any particular intermediate state can be used to isolate a graph subspace that can be approached from that state by applying the grammar rules. Therefore, one can use the finished mechanisms in the subspace for intermediate state evaluation.

\subsection{Graph search algorithm} 
\label{MCTS}
To explore the vast design space of possible solutions we need a heuristic search algorithm. Currently, the applied algorithms support only generation of open chain mechanisms. Therefore, we can restrict grammar rules so that they only add elements to design morphology. The search space can be represented as a tree with all possible states. We can use any search algorithm to find solutions on the tree. 

\textit{Monte-Carlo Tree Search} (MCTS) is well-suited for exploring the graph grammar-generated search space. The graph $g \in \mathcal{G}$ of the mechanism is a leaf in the search tree. The directed edges represent possible rules~$\mathcal{R}^v$ from the rule dictionary~$\mathcal{R}$. MCTS initiate the reward computation only for the leaf graphs in order to estimate the $\hat V$ value function of the states. 

The MCTS has only four steps: selection, expansion, simulation, and back propagation. During selection, the algorithm looks for a leaf node $\aleph$ in the search tree. It applies the Upper Confidence Bound (UCB) method in each node until it reaches the leaf.
The bandit's UCB formula is:
$$
r := \argmax_{r \in \mathcal{R}^v} \left [Q(\aleph,r) + C\sqrt{\frac{\ln{n(\aleph)}}{ n(\aleph,r)}} \right ]
$$
where $Q(\aleph,r)$ is the function of the maximum possible reward when rule $r$ is applied in node $\aleph$, $C$ is the hyperparameter, $n(\aleph)$ is the count of visiting node $\aleph$, $n(\aleph, r)$ is the count of apply ing rule $r$ in $\aleph$. Addition of UCB bandit to MCTS is called the UCT algorithm~\cite{mcts_uct}.

In the expansion step, the algorithm applies valid rules $r \in \mathcal{R}^v$ to a current state. In the simulation step of the MCTS the estimation of child nodes rewards $\hat \aleph$ is occurred. The algorithm applies randomly selected rules until it reaches the terminal state. The special parameter restricts the application of rules, to guarantee that the algorithm always gets to the terminal state in finite number of steps. During backpropagation, the state-value function $Q(s,a)$ rules and node on the trajectory are updated. 

We implemented the MCTS algorithm because it is simple to implement, it showed acceptable results, and it can be used to compare other more efficient search algorithms, such as GNNs~\cite{zhao2020robogrammar}, that we plan to integrate later. 

\section{Generation of tendon driven grippers}

%The particular task for the GD should be set to get the results. 
%In this section we show an application how the proposed design procedure can be used for the task of generation of tendon-driven anthropomorphic grippers. %Here we focus on the morphological aspect of mechanisms, 
%hardly restricted other design choices. 
%We created production rules, an actuation scheme, and topology components to easily test our prototype. Having in mind the \textit{reality gap}, we tried to choose the appropriate reward function. 

\begin{figure}[b!] 
	\begin{center}
		\includegraphics[width=0.99\linewidth]{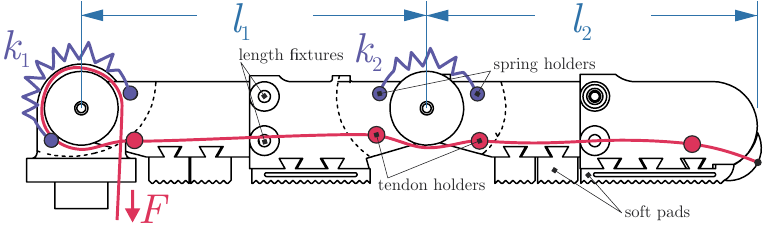}
		\caption{Schematic representation of a finger with two phalanges in a horizontal state. }
		% each finger is equiped with a prizamtic joint 
		\label{fig:finger}
	\end{center}
\end{figure}

\subsection{Finger design}
%According to the general scheme, first step is to determine what role would play the graph search in the whole GD process.
%We limit the optimization of parameters of robot components to a choice from several discrete values and in this form add it to the graph search task. On the other hand, the control optimization is set to be carried out independently as a part of the reward calculation.  
We simplify the process of optimizing robot component parameters by selecting from a few specific values. The grippers we design consists of the palm and up to four fingers where each finger can have up to five phalanges. The palm serves as the central point of a star topology graph, with each finger branching out as a separate connection. This representation encompasses all mechanisms of a similar nature. Graph grammar rules consist of nodes and edges. The nodes contain all the information, while the edges determine the sequence of details. As a result, each finger consists of three types of nodes: \textit{phalanxes, joints} and \textit{transforms} along the palm.
%The length of each phalanx, stiffness of each joint and the position of a finger on a palm can be chosen from the predetermined discrete set, that must be given by a framework user.

Fig.~\ref{fig:finger} shows schematic representation of a finger with two phalanges. For each finger algorithm discretely changes base's position and orientation on a palm, links' number and their lengths $l_i$, spring stiffness $k_i$, and driving force $F$. The upper and lower bounds for the parameters were manually selected based on the sizes of the objects to be grasped and the components available in the lab, i.e. spring and actuators. We kept the pulley radius constant to simplify the procedure and focus on verifying the general method instead of finding an exact solution.  
%We then incorporate this simplified process into the graph search task. On the other hand, The control optimization is set to be carried out independently as a part of the reward calculation.  

%It is crucial to define the graph grammar and establish the necessary limitations.

\subsection{Production rules and robot components} 
\label{rules}
The whole set of production rules is illustrated in Fig.~\ref{fig:pic_2}. We use \textit{terminal} and \textit{non-terminal nodes }in our concept. Non-terminal nodes represent the abstract component which only provides information about the robot's structure, while terminal nodes represent components with specific physical properties. 

The terminal symbols on the scheme represent subsets of terminal nodes. These subsets are used to change the corresponding parameters. For example, the terminal L represents a set of discrete links' lengths, where each terminal node in the set has its own phalanx size. This graph grammar has two types of rules: (1) type one -- changes design morphology, and (2) type two -- transforms non-terminal nodes into terminal ones, determining the design parameters.    

Even such a simple set allows us to generate vast space of the possible designs, which is another advantage of the graph grammar approach. The design space includes common designs like symmetric two and three finger grippers. It also allows for more unique variations, such as three fingers with different shifts and rotations on one side of the palm, and an opposing finger on the other side. Fig.~\ref{fig:pic_3} shows illustrative example of a sequence of rules that can be applied to generate the simplest one-fingered gripper. 
%Figure illustrates  

\subsection{Virtual experiment setup}

% In a tendon-driven mechanism, tendons are used to transmit forces from an actuator, like a motor, to a load or output. The tendons are typically attached to pulleys or other mechanisms that change the direction of the force or provide mechanical advantages~\cite{wang2011highly}. 

%Each finger is independently actuated through a tendon. Each phalanx has two points of force application and force vector in each point depends only on the current configuration of the finger and the tendon tension provided by the virtual motor~\cite{wang2011highly}. The relative positions of virtual pulleys in the phalanx are constant, thus the only parameter for control optimization is the force applied to the tendon. We simplified the task by focusing on design morphology and algorithm testing. We only used three values for each finger: \textit{weak, intermediate,} and \textit{strong} tension. As a result, each mechanism has a control parameter space with $3^n$ points, where $n$ is the number of fingers.  

%We model the situation where the thorough analysis of the object or the perception are impossible. Consider that the only accessible information is the rough size of the object. %(the size of a sphere that is guaranteed to cover the object). 
%In the design evaluation loop, 

The algorithm simulates the gripper's interaction with various objects at different control (torque value) parameter settings. Within this study we fixed the dimension of the objects range to three, thus, the evaluation of each mechanism includes $3 \cdot 3^n$ simulations. A simulation consists of two stages. At first stage the gripper attempts to grasp an object located in the center of the gripper workspace. To grasp objects blindly, we need to be able to hold them regardless of their initial orientation. To reduce its effect, we turn off gravity so that grasping is done only by using encompassing feature grippers, rather than relying on the force of gravity. At the second stage, if the grasp is successful, the algorithm applied an external load proportional to object weight. If the object remains grasped and does not fly away, then the grasp is considered successful.

\begin{figure*}[t] 
	\begin{center}
		\includegraphics[width=0.90\textwidth]{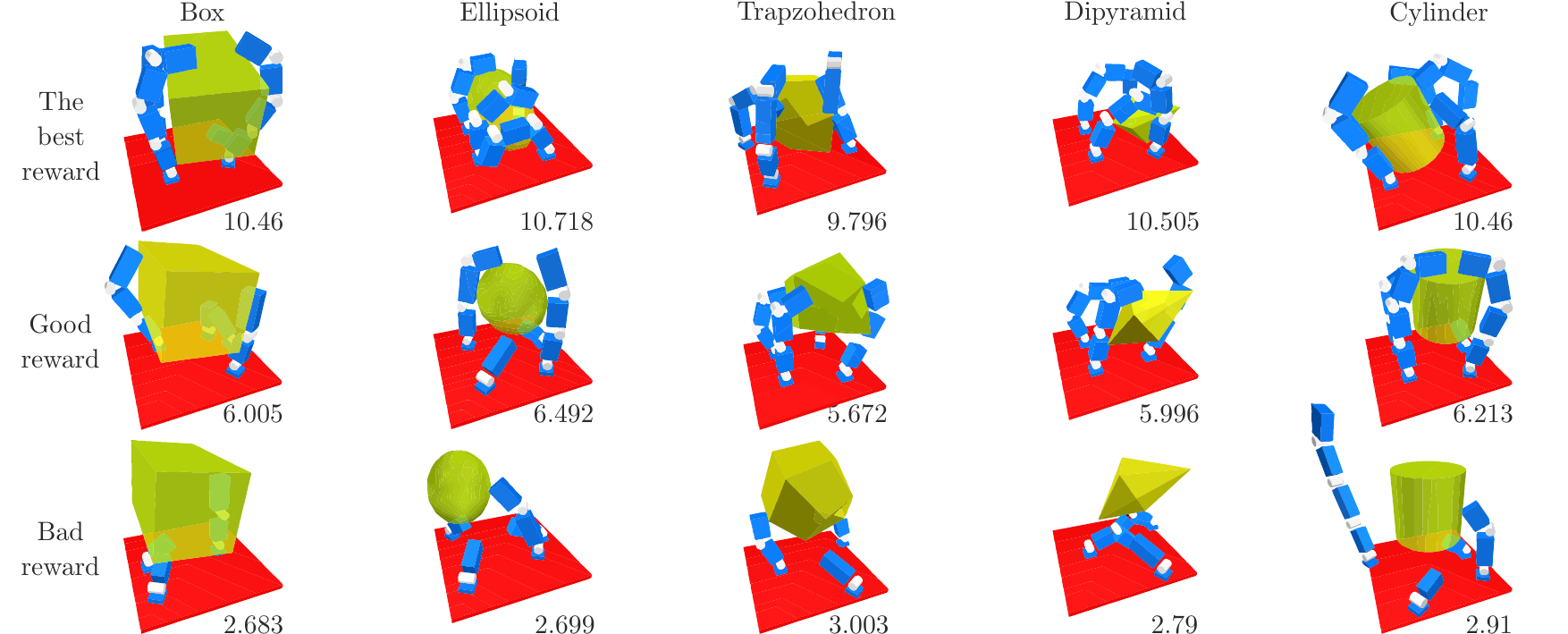}
		\caption{Samples showing simulation models of generated gripper designs. Numbers next to images are values of a reward function used to evaluate the whole performance of generated graphs. Values around 10 indicate secure grasp of an object even with external force applied, values around 6 mean that an object was grasped, but was not able to handle an external force, and finally rewards around 3 means gripper have just touched an object without grasping} 
		\label{fig:pic_4}
	\end{center}
\end{figure*}

\subsection{Reward function}

We designed several quantitative criteria to evaluate the performance of a mechanism in a simulation. Here we present a short description of each criterion.

\subsubsection{Time period for the grasp} we abort the simulation prematurely if the mechanism cannot touch an object or the contact is lost. The simulation time can be used as a criterion: 

% We abort the simulation early if the mechanism can't touch the object or loses contact with it. We use the length of time the mechanism stays in contact with the object as a measure of success:

$$r_1 = {1}/(1+t_{sim}^2/t_{max}^2),$$
where $t_{sim} \in \mathbb{R}$ is the current
% where $t_{sim} \in \mathbb{R}$ is elapsed simulation time 
time, $t_{max}\in \mathbb{R}$ is the limited simulation time.
%%%%%%%%%%%%%%
The criterion distinguishes unsuccessful designs, meaning that achieving longer contact with an object is preferable. Designs that do not touch the object receive zero reward. All mechanisms that successfully hold the object receive a score of~3~for~$r_1$. For example, bad solutions depicted in Fig.~\ref{fig:pic_4}) have reward value less than 3 meaning that fingers only have touched the objects without grasping. That criterion is calculated when the object is grasped just before the external force to be applied. 

\subsubsection{The fraction of phalanxes contacting with objects} needed to reward for effective use of the phalanxes to prevent the growth of unnecessary components:
$$ r_2 = {\sum_{b \in \mathcal{B}} \mathbb{I}(b)}/{||\mathcal{B}||},$$ where
$\mathcal{B} \in \mathbb{N}$ is a set of all mechanism bodies in simulation and $\mathbb{I}$ is an indicator function that equals to 1 if the corresponding body contacts an object; thus, the numerator of $r_2$ gets plus one for each body in contact with the object. 
%%%
\subsubsection{Dispersion of the contact forces} for a secure and stable grasp, phalanges have to apply similar forces to an object to be grasped 
%The grasp that rely on applying much force on specific point could damage an object or a corresponding element of the gripper, while the weak contact cannot withstand the external force:
%$$r_3 = 1/[1 + \mathrm{std}(\mathbf{f}(t_{\mathrm{grasp}}))],$$ 
$$r_3 = 1/[1 + \mathrm{std}(||f_i(t_{\mathrm{grasp}})||)],$$
where $f(t_\mathrm{grasp})$ is the force vector applied by a generated mechanism to a grasped object, \text{std} stands for standard deviation, $t_{\mathrm{grasp}}$ is time needed to secure the object. 
%%%

\subsubsection{Distance between an object center and a geometric center of contact points}
the small distance reduce inertia effects during the grasping and helps to achieve a stable grasp:
$$
r_4 = 1/( 1 +  {\\||g_i - c_i\\||}_2 ),
$$
where $g_i\in \mathbb{R}^3$ is a geometric center of contact points and $c_i \in \mathbb{R}^3$ is a center of gravity.

\subsubsection{Grasp time} time needed to secure the grasped object. Quick grasp gets higher reward: 
$$ r_5~=~(t_\mathrm{max} - t_\mathrm{grasp})/t_\mathrm{max}.$$

\subsubsection{The ability to withstand external force} when the grasp is secured, algorithm applies gradually increasing force to the grasped object. The force is proportional to the corresponding weight. Zero reward is given if the external force takes the object away from the gripper. If the gripper holds the grasped object, it gets the following reward:
$$
r_6 = 1 - {p}(t_\mathrm{final}) / {p}(t_\mathrm{grasp}),
$$ where ${p}(t) \in \mathbb{R}^3$ is a position vector of a grasped object, $t_\mathrm{final}$ is the final time of the simulation.

\subsubsection{Final reward} 
%%%
%The reward that a design get in a simulation is a weighted sum of described partial rewards.  
A generated gripper gets a scalar reward after each simulation, and we use its value to evaluate the design candidate:
$$ r = w_1 r_1 + w_2 r_2 + w_3 r_3 + w_4 r_4 + w_5 r_5 + w_6  r_6,$$
where the weight coefficients $w_i$ are equal to 1, except $w_1=3$, and $w_6=5$, because of higher priority. 
%%%
In Fig.~\ref{fig:pic_4} samples are given. For example, a reward decomposition for the gripper with the best reward that holds a box is the following
$$ r = 3 \cdot 1 + 2 \cdot 0.31 + 1 \cdot 0.36 + 1 \cdot 0.96 + 1 \cdot 0.77 + 5  \cdot 0.95 = 10.46.$$

We determine the highest value in the control space for each object we grasp. The sum of the highest rewards for all tested objects is then selected as the final reward for the design. Finally, the constructed reward function is used to search the design space applying the MCTS algorithm described in \ref{MCTS}.

\subsubsection{Remark on reward, bounds and initial state} 

\begin{figure}[b!] 
	\begin{center}
		\includegraphics[width=0.85\linewidth]{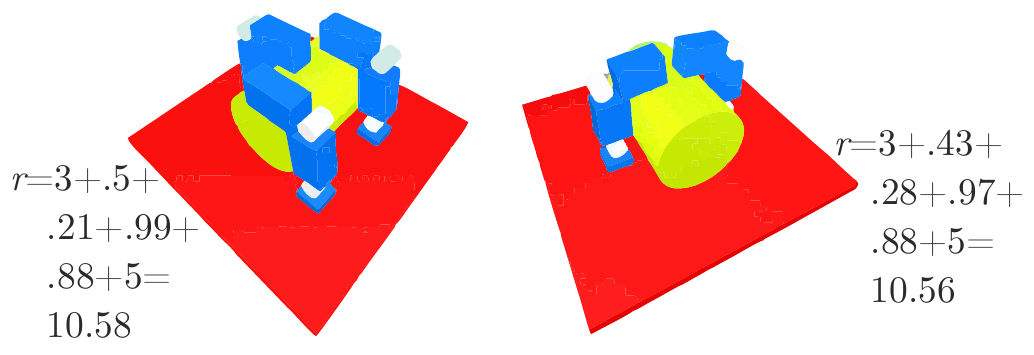}
		\caption{Generic grippers found with corresponding topology and parameters limits together with choosing a different set of weight coefficients}
		% each finger is equiped with a prizamtic joint 
		\label{fig:generic}
	\end{center}
\end{figure}

The weight coefficients in the reward function have a substantial impact on the final designs, along with the initial orientation of objects and the limits set on the structure and parameters. If we limit the number of fingers to the conventional number of three together with limiting the number of phalanges we are getting generic grippers depicted in Fig.~\ref{fig:generic}. From this sense, the main advantage of the proposed procedure is that the algorithm is capable of searching through the vast design space, and \textit{what} will be the best solution depends on the reward.

\begin{figure*}[t] 
	
	\begin{center}
		\includegraphics[width=0.99\textwidth]{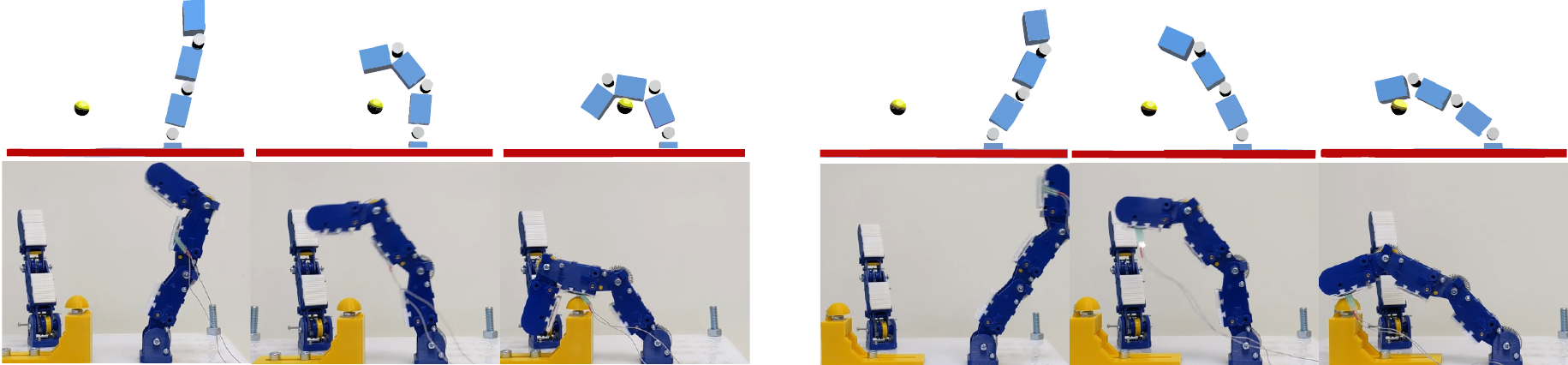}
		\caption{Comparison between sequences of motion of a simulation model of one finger and corresponding physical prototype. To measure contact forces a penalty based method has been used in simulation, while for the physical prototype we have used a force resistive sensor} 
		\label{fig:pic_6}
	\end{center}
	\begin{center}
		\includegraphics[width=0.99\textwidth]{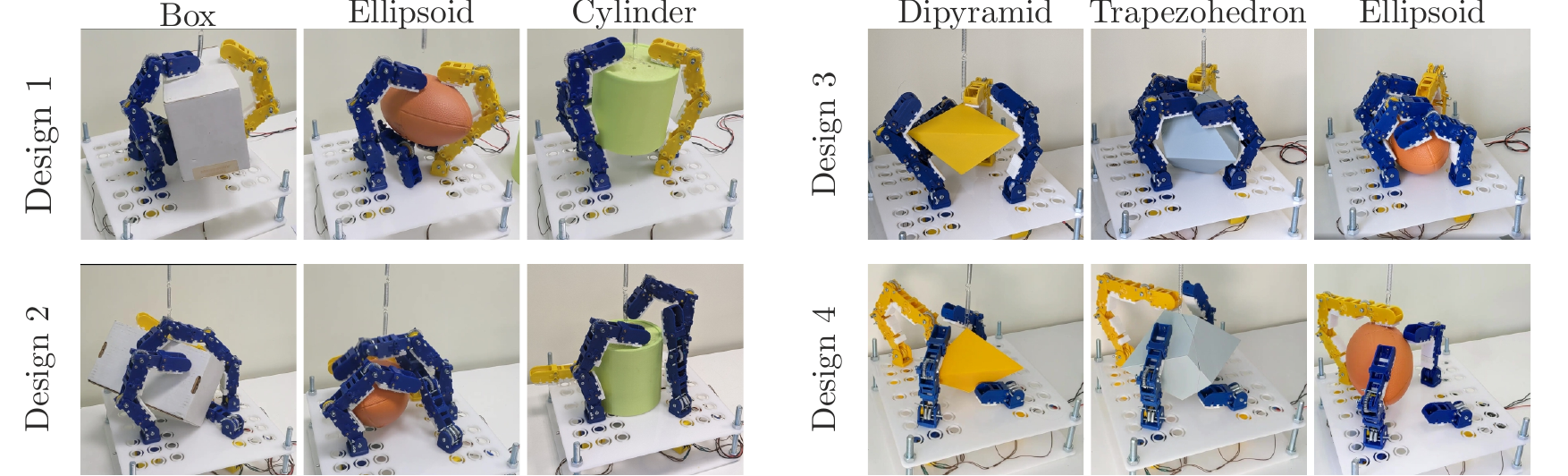}
		\caption{Visualization of manufacturing physical grippers according to the generated designs with the top of reward function. Grasping is primarily done because of devices' morphology, rather than sophisticated control. Shown designs are capable of securely grasping different objects even with external~force} 
		\label{fig:pic_7}
	\end{center}
	%\end{figure*}
	% \begin{figure*}[t] 
		
	\end{figure*}

% physacal design is needed to specifiy parameters such that they make sense 

\section{Results discussion}

\subsection{Search results}
\begin{figure}[b!] 
	\begin{center}
		\includegraphics[width=0.8\linewidth]{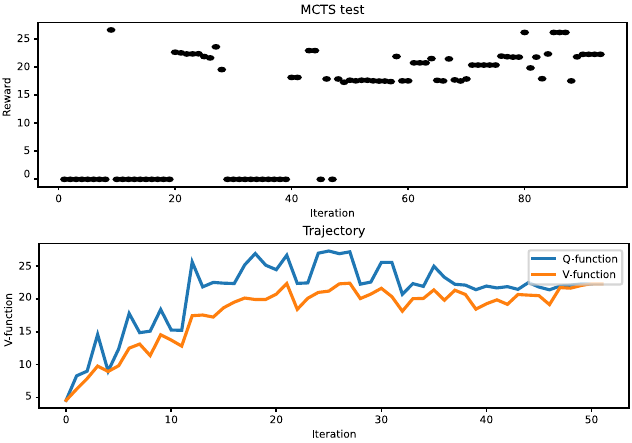}
		\caption{Relation between reward values and a number of iterations together with plots for Q and V function for MCTS heuristic search algorithm} 
		\label{fig:pic_5}
	\end{center}
\end{figure}

The algorithm evaluates and saves encoded mechanism representations and rewards during the design search process. We consider the designs with top rewards %(or several designs with top rewards)
as a set of final results. % of the GD algorithm. 
The designs generated are shown in Fig.~\ref{fig:pic_4}. The quantitative characteristics of the MCTS run are shown in Fig.~\ref{fig:pic_5}. The increasing V and Q functions demonstrate that the search algorithm consistently achieves higher average rewards over time. %As a result, 

The search space expands to include more valuable designs. A test run, that is performed in each step illustrates the current learned model of the search space. After 40 iterations, the reward increases significantly. Using the learned design space model, the MCTS constructed a decision tree that extended to terminal states. It then proceeded to choose rules based on this tree. 
Once the learned MCTS has gone through a significant number of iterations, it constructs a decision tree that leads to terminal states with a highly promising outcome. Simulation took 2 days on a machine with CPU i9 10920X with 12 cores to conduct all calculations.

\subsection{Experimental setup}

We created a physical setup to validate our approach and verify the results~(Fig.~\ref{fig:pic_7}). The CAD model was designed with consideration of the graph grammar described in \ref{rules}, which ensures that any generated graph could be tested in the physical world in a fast and efficient manner. The phalanx was chosen as the foundational building block of the prototype, and it was designed to be adjustable. Each link is equipped with a prismatic joint (Fig.~\ref{fig:finger}), such that its length can be varied. This versatility allowed us to assemble various configurations with different numbers of fingers, number of phalanxes and lengths. The design includes features from the tendon-driven mechanism such as pulleys for directing the tendons 
and springs. The position of the pulleys and the stiffness of the springs were selected to resemble the virtual experiments for more credible validation of our methodology. 

Additionally, the palm component was specifically designed to facilitate testing across all finger positions and orientations that our method can produce. 
In terms of manufacturing, we employed 3D printing for the phalanges and laser cutting for the palm for %. These choices were made to accomplish 
%rapid (re-)assembling of different prototypes, while simultaneously ensuring structural integrity and high build quality. 
The Dynamixel AX-12A servo were chosen to control the fingers due to their suitable torque characteristics and diverse feedback capabilities. We added FSR402 sensors to measure grasping pressure. To manage and communicate with these sensors and motors, we used an STM32F4 microcontroller.

\subsection{Verification}

The tendon-driven finger is an underactuated open chain mechanism, thus its % FIX move pattern
movement pattern and contact forces depend on both tendon tension and pulley positions. In order to verify the precision of the testing setup, we carried out a set of experiments with a singular finger. The testing setup is illustrated in Fig.~\ref{fig:pic_6}. Motion of the physical finger has been % FIX compered 
compared with the virtual one by visual inspection. We measured the contact force between the test finger and the fixed object using the FSR402 sensor to determine the difference between the simulation and the experiment. The difference in simulated and experimental contact forces in our tests did not exceed 34\% (Fig.~\ref{fig:pic_6}). 

It was noticed, that a larger number of fingers have been generated on the side where the external force has been applied. The grasp is more secured if the fingers wrap around from different directions like tentacles. Thus, as for the grippers on Fig.~\ref{fig:pic_7} we have analyzed the grasping patters visually. Because of underactuation, a significant effect on object initial state, and "flying" objects the motion of the fingers differs from the simulation. Nevertheless, because we have trained them on a set of objects with randomly applied forces, the grippers are versatile enough to encompass the objects even if they are oriented differently. 

\section{Conclusion}
This paper explores the intricate process of designing robots, emphasizing the convergence of hardware and software in complex, conflicting criteria. We introduced a novel approach combining \textit{morphological computation}, according to which the major portion of robots’ desired
behavior can be achieved with the “body” instead of the
“brain”, and \textit{generative design}, highlighting the potential of automatic design for underactuated tendon-driven grippers, with open-source framework details and comprehensive testing results.
The result of the ``rostok"\footnote{All methods and algorithms described in the paper are available as a part of the open-source framework \textit{rostok} (\url{https://github.com/aimclub/rostok})} pipeline for the task of generating tendon driven grippers is the set of designs with top rewards. The designs were thoroughly tested in a physical setup to ensure their ability to accurately reproduce the simulated kinematics and securely hold the object in place. 

%As a result of the look over the design space we get a set of designs and the corresponding set of rewards. To verify our approach to generative design we built several mechanisms obtained in the search process and compared their behavior in the real experiment.   

%\section{Data and code availability}
%All methods and algorithms described in the paper are available as a part of the open-source library \textit{rostok} (\url{https://github.com/aimclub/rostok}).
%\section{Data and code availability}
%All methods and algorithms described in the paper are available as a part of the open-source library \textit{rostok} (\url{https://github.com/aimclub/rostok}).

\bibliographystyle{IEEEtran}
\bibliography{co_design_open} 

\begin{thebibliography}{10}
\providecommand{\url}[1]{#1}
\csname url@samestyle\endcsname
\providecommand{\newblock}{\relax}
\providecommand{\bibinfo}[2]{#2}
\providecommand{\BIBentrySTDinterwordspacing}{\spaceskip=0pt\relax}
\providecommand{\BIBentryALTinterwordstretchfactor}{4}
\providecommand{\BIBentryALTinterwordspacing}{\spaceskip=\fontdimen2\font plus
\BIBentryALTinterwordstretchfactor\fontdimen3\font minus \fontdimen4\font\relax}
\providecommand{\BIBforeignlanguage}[2]{{%
\expandafter\ifx\csname l@#1\endcsname\relax
\typeout{** WARNING: IEEEtran.bst: No hyphenation pattern has been}%
\typeout{** loaded for the language `#1'. Using the pattern for}%
\typeout{** the default language instead.}%
\else
\language=\csname l@#1\endcsname
\fi
#2}}
\providecommand{\BIBdecl}{\relax}
\BIBdecl

\bibitem{sun2021research}
Y.~Sun, J.~Falco, M.~A. Roa, and B.~Calli, ``Research challenges and progress in robotic grasping and manipulation competitions,'' \emph{IEEE robotics and automation letters}, vol.~7, no.~2, pp. 874--881, 2021.

\bibitem{borisov2021computational}
I.~I. Borisov, E.~E. Khomutov, S.~A. Kolyubin, and S.~Stramigioli, ``Computational design of reconfigurable underactuated linkages for adaptive grippers,'' in \emph{2021 IEEE/RSJ International Conference on Intelligent Robots and Systems (IROS)}.\hskip 1em plus 0.5em minus 0.4em\relax IEEE, 2021, pp. 6117--6123.

\bibitem{borisov2022reconfigurable}
I.~I. Borisov, E.~E. Khornutov, D.~V. Ivolga, N.~A. Molchanov, I.~A. Maksimov, and S.~A. Kolyubin, ``Reconfigurable underactuated adaptive gripper designed by morphological computation,'' in \emph{2022 International Conference on Robotics and Automation (ICRA)}.\hskip 1em plus 0.5em minus 0.4em\relax IEEE, 2022, pp. 1130--1136.

\bibitem{nasonov2023computational}
K.~V. Nasonov, D.~V. Ivolga, I.~I. Borisov, and S.~A. Kolyubin, ``Computational design of closed-chain linkages: Hopping robot driven by morphological computation,'' in \emph{2023 IEEE International Conference on Robotics and Automation (ICRA)}.\hskip 1em plus 0.5em minus 0.4em\relax IEEE, 2023, pp. 7419--7425.

\bibitem{thomaszewski2014computational}
B.~Thomaszewski, S.~Coros, D.~Gauge, V.~Megaro, E.~Grinspun, and M.~Gross, ``Computational design of linkage-based characters,'' \emph{ACM Transactions on Graphics (TOG)}, vol.~33, no.~4, pp. 1--9, 2014.

\bibitem{de2021control}
F.~De~Vincenti, D.~Kang, and S.~Coros, ``Control-aware design optimization for bio-inspired quadruped robots,'' in \emph{2021 IEEE/RSJ International Conference on Intelligent Robots and Systems (IROS)}.\hskip 1em plus 0.5em minus 0.4em\relax IEEE, 2021, pp. 1354--1361.

\bibitem{chen2020underactuation}
T.~Chen, L.~Wang, M.~Haas-Heger, and M.~Ciocarlie, ``Underactuation design for tendon-driven hands via optimization of mechanically realizable manifolds in posture and torque spaces,'' \emph{IEEE Transactions on Robotics}, vol.~36, no.~3, pp. 708--723, 2020.

\bibitem{bhatia2021evolution}
J.~Bhatia, H.~Jackson, Y.~Tian, J.~Xu, and W.~Matusik, ``Evolution gym: A large-scale benchmark for evolving soft robots,'' \emph{Advances in Neural Information Processing Systems}, vol.~34, 2021.

\bibitem{cheney2018scalable}
N.~Cheney, J.~Bongard, V.~SunSpiral, and H.~Lipson, ``Scalable co-optimization of morphology and control in embodied machines,'' \emph{Journal of The Royal Society Interface}, vol.~15, no. 143, p. 20170937, 2018.

\bibitem{ha2021fit2form}
H.~Ha, S.~Agrawal, and S.~Song, ``Fit2form: 3d generative model for robot gripper form design,'' in \emph{Conference on Robot Learning}.\hskip 1em plus 0.5em minus 0.4em\relax PMLR, 2021, pp. 176--187.

\bibitem{hu2022modular}
J.~Hu, J.~Whitman, M.~Travers, and H.~Choset, ``Modular robot design optimization with generative adversarial networks,'' in \emph{2022 International Conference on Robotics and Automation (ICRA)}.\hskip 1em plus 0.5em minus 0.4em\relax IEEE, 2022, pp. 4282--4288.

\bibitem{zhao2020robogrammar}
A.~Zhao, J.~Xu, M.~Konakovi{\'c}-Lukovi{\'c}, J.~Hughes, A.~Spielberg, D.~Rus, and W.~Matusik, ``Robogrammar: graph grammar for terrain-optimized robot design,'' \emph{ACM Transactions on Graphics (TOG)}, vol.~39, no.~6, pp. 1--16, 2020.

\bibitem{wang2019neural}
T.~Wang, Y.~Zhou, S.~Fidler, and J.~Ba, ``Neural graph evolution: Towards efficient automatic robot design,'' \emph{arXiv preprint arXiv:1906.05370}, 2019.

\bibitem{gupta2022metamorph}
A.~Gupta, L.~Fan, S.~Ganguli, and L.~Fei-Fei, ``Metamorph: Learning universal controllers with transformers,'' \emph{arXiv preprint arXiv:2203.11931}, 2022.

\bibitem{pathak19assemblies}
D.~Pathak, C.~Lu, T.~Darrell, P.~Isola, and A.~A. Efros, ``Learning to control self- assembling morphologies: A study of generalization via modularity,'' in \emph{NeurIPS}, 2019.

\bibitem{zlokapa2022integrated}
L.~Zlokapa, Y.~Luo, J.~Xu, M.~Foshey, K.~Wu, P.~Agrawal, and W.~Matusik, ``An integrated design pipeline for tactile sensing robotic manipulators,'' in \emph{2022 International Conference on Robotics and Automation (ICRA)}.\hskip 1em plus 0.5em minus 0.4em\relax IEEE, 2022, pp. 3136--3142.

\bibitem{rocchi2016generic}
A.~Rocchi and K.~Hauser, ``A generic simulator for underactuated compliant hands,'' in \emph{2016 IEEE International Conference on Simulation, Modeling, and Programming for Autonomous Robots (SIMPAR)}.\hskip 1em plus 0.5em minus 0.4em\relax IEEE, 2016, pp. 37--42.

\bibitem{wang2018neural}
\BIBentryALTinterwordspacing
T.~Wang, Y.~Zhou, S.~Fidler, and J.~Ba, ``Neural graph evolution: Automatic robot design,'' in \emph{International Conference on Learning Representations}, 2019. [Online]. Available: \url{https://openreview.net/forum?id=BkgWHnR5tm}
\BIBentrySTDinterwordspacing

\bibitem{Sims_1994}
K.~Sims, ``Evolving virtual creatures,'' \emph{Proceedings of the 21st annual conference on Computer graphics and interactive techniques nbsp; - SIGGRAPH ’94}, 1994.

\bibitem{tasora2016chrono}
A.~Tasora, R.~Serban, H.~Mazhar, A.~Pazouki, D.~Melanz, J.~Fleischmann, M.~Taylor, H.~Sugiyama, and D.~Negrut, ``Chrono: An open source multi-physics dynamics engine,'' in \emph{High Performance Computing in Science and Engineering: Second International Conference, HPCSE 2015, Sol{\'a}{\v{n}}, Czech Republic, May 25-28, 2015, Revised Selected Papers 2}.\hskip 1em plus 0.5em minus 0.4em\relax Springer, 2016, pp. 19--49.

\bibitem{mcts_uct}
L.~Kocsis and C.~Szepesv{\'a}ri, ``Bandit based monte-carlo planning,'' in \emph{Machine Learning: ECML 2006: 17th European Conference on Machine Learning Berlin, Germany, September 18-22, 2006 Proceedings 17}.\hskip 1em plus 0.5em minus 0.4em\relax Springer, 2006, pp. 282--293.

\end{thebibliography}

\end{document}